\documentclass[conference]{IEEEtran}
\IEEEoverridecommandlockouts
\usepackage{cite}
\usepackage{amsmath,amssymb,amsfonts}
\usepackage{graphicx}
\usepackage{textcomp}
\usepackage{xcolor}
\def\BibTeX{{\rm B\kern-.05em{\sc i\kern-.025em b}\kern-.08em
    T\kern-.1667em\lower.7ex\hbox{E}\kern-.125emX}}

\usepackage{algorithm}
\usepackage{algpseudocode}
\usepackage{multirow}
\usepackage{booktabs}
\usepackage{bm}
\usepackage{paralist}
\usepackage{enumitem}
\usepackage{bbding}

\newtheorem{thm}{Theorem}
\newtheorem{lm}{Lemma}

\newtheorem{as}{Assumption}

\begin{document}

\title{Communication-Efficient Training Workload Balancing for Decentralized Multi-Agent Learning}

\author{\hspace{-1.5cm}\IEEEauthorblockN{Seyed Mahmoud Sajjadi Mohammadabadi }
\IEEEauthorblockA{\hspace{-1.5cm}\textit{Department of Computer Science and Engineering} \\
\hspace{-1.5cm}\textit{University of Nevada, Reno}\\
\hspace{-1.5cm} Reno, NV, USA \\
\hspace{-1.5cm} mahmoud.sajjadi@unr.edu}
\and
\hspace{-1cm}\IEEEauthorblockN{Lei Yang}
\IEEEauthorblockA{\hspace{-0cm}\textit{Department of Computer Science and Engineering}\hspace{1cm} \\
\hspace{-1cm}\textit{University of Nevada, Reno}\\
\hspace{-1cm}Reno, NV, USA \\
\hspace{-1cm}leiy@unr.edu}
\and
\hspace{2cm}\IEEEauthorblockN{Feng Yan}
\IEEEauthorblockA{\hspace{2cm}\textit{Department of Computer Science} \\
\hspace{2cm}\textit{University of Houston}\\
\hspace{2cm}Houston, TX, USA \\
\hspace{2cm}fyan5@central.uh.edu}
\and
\hspace{2cm}\IEEEauthorblockN{Junshan Zhang}
\IEEEauthorblockA{\hspace{2cm}\textit{Department of Electrical and Computer Engineering} \\
\hspace{2cm}\textit{University of California, Davis}\\
\hspace{2cm}Davis, CA, USA \\
\hspace{2cm}jazh@ucdavis.edu}
}

\maketitle

\begin{abstract}
Decentralized Multi-agent Learning (DML) enables collaborative model training while preserving data privacy. However, inherent heterogeneity in agents' resources (computation, communication, and task size) may lead to substantial variations in training time. This heterogeneity creates a bottleneck, lengthening the overall training time due to straggler effects and potentially wasting spare resources of faster agents.
To minimize training time in heterogeneous environments, we present a Communication-Efficient Training Workload Balancing for Decentralized Multi-Agent Learning (ComDML), which balances the workload among agents through a decentralized approach. Leveraging local-loss split training, ComDML enables parallel updates, where slower agents offload part of their workload to faster agents. To minimize the overall training time, ComDML optimizes the workload balancing by jointly considering the communication and computation capacities of agents, which hinges upon integer programming. A dynamic decentralized pairing scheduler is developed to efficiently pair agents and determine optimal offloading amounts.
We prove that in ComDML, both slower and faster agents' models converge, for convex and non-convex functions.
Furthermore, extensive experimental results on popular datasets (CIFAR-10, CIFAR-100, and CINIC-10) and their non-I.I.D. variants, with large models such as ResNet-56 and ResNet-110, demonstrate that ComDML can significantly reduce the overall training time while maintaining model accuracy, compared to state-of-the-art methods. ComDML demonstrates robustness in heterogeneous environments, and privacy measures can be seamlessly integrated for enhanced data protection.

\end{abstract}


\section{Introduction}

Effective training of Deep Neural Networks (DNNs) often requires access to a vast amount of data typically unavailable on a single device. Transferring data from different devices (a.k.a., clients or agents) to a central server for training raises security and privacy concerns, as well as communication costs and challenges. To address these issues, there is a growing trend towards cooperative training of machine learning models across a network of devices, eliminating the need to transfer the local training data. Federated Learning (FL) \cite{mcmahan2017communication} algorithms have gained substantial attention as a privacy-preserving distributed learning paradigm. In FL, a central server acts as a coordinator among participating agents, enabling them to update a global model using their locally trained weights.
The training process of FL, however, causes a major challenge when dealing with real-world resource-constrained devices (e.g., mobile/IoT devices and edge servers) that often exhibit heterogeneous computation and communication capacities, along with varying dataset sizes. 
Such heterogeneity not only introduces substantial variations in training time across agents, leading to the straggler problem (i.e., some devices significantly lag behind others) but also wastes the available spare resources of faster agents.

To address the challenges due to the unbalanced workload on resource-constrained devices, different methods have been proposed recently. One popular approach involves splitting the global model into an agent-side model (consisting of the initial layers of a global model) and a server-side model (the remaining layers), where agents only need to train the smaller agent-side model using Split Learning (SL) \cite{vepakomma2018split,thapa2022splitfed}. However, SL requires agents to wait for backpropagated gradients from the server to update their models, resulting in substantial communication overhead in each training round. 
To address the latency and communication issues of SL, a federated SL algorithm is developed by incorporating local-loss-based training \cite{han2021accelerating}. However, their approach uses fixed agent-side models, limiting their adaptability to varying computation and communication resources in dynamic environments.
Along another line, agents can be segmented into tiers based on their training speed, and agents from the same tier are selected in each training round to mitigate the straggler problem \cite{chai2020tifl,fedatSC21}. However, existing tier-based approaches \cite{chai2020tifl,fedatSC21} require agents to train the entire global model locally, which is not scalable for training large models. 
For these methods \cite{vepakomma2018split,han2021accelerating, chai2020tifl, fedatSC21, thapa2022splitfed}, a central server is required to coordinate the training of all agents.
A centralized server, both prone to latency bottlenecks \cite{beltran2023decentralized, zong2024fedcs} and susceptible to failures and targeted attacks \cite{ma2020safeguarding}, can significantly undermine the reliability of the entire distributed learning process.
To mitigate these issues, decentralized (peer-to-peer) systems have emerged as an alternative \cite{roy2019braintorrent, hegedHus2019gossip, tang2023gossipfl}. 
Distinct from distributed systems that utilize a central server for coordination, these systems rely on peer-to-peer communication, offering improved resilience and security due to the absence of a single point of failure. However, without the coordination of a centralized scheduler, workload balancing in these decentralized systems becomes challenging.

\begin{figure}[t]
  \centering
   \includegraphics[width=1.00\linewidth]{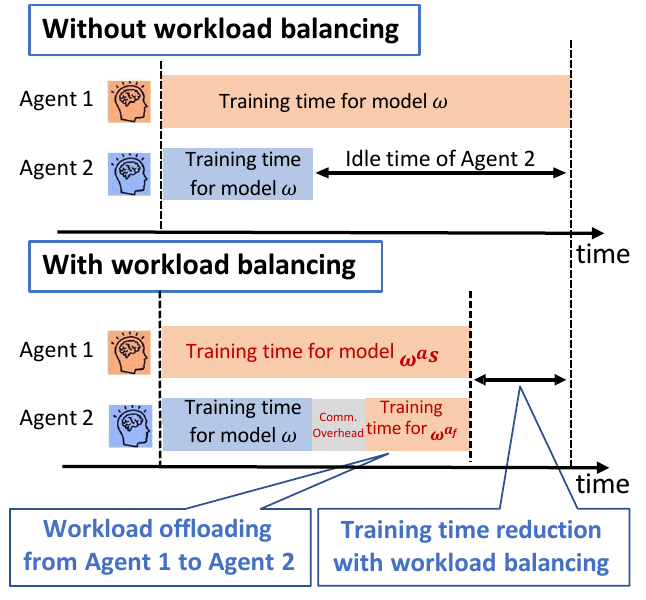}

   \caption{Comparison of model training with and without workload balancing. Workload balancing reduces training time by offloading the workload from agent 1 to agent 2, which would otherwise be idle.}
   \label{fig:workload_balancing}
\end{figure}

In this paper, we propose a novel Communication-Efficient Training Workload Balancing for Decentralized Multi-Agent Learning (ComDML) that effectively addresses the challenges of training workload balancing in decentralized systems, operating without a server or coordinator.
In ComDML, the training workload is balanced by allowing slower agents to offload a portion of their workload to faster agents, ensuring efficient utilization of available resources (see Fig. \ref{fig:workload_balancing}). 
To reduce synchronization and communication overhead between paired agents, ComDML employs local-loss-based split training, where the paired agents can determine how to split the model and then train the split model in parallel.
ComDML's core objective is to minimize the overall training time. ComDML achieves it by employing an integer programming formulation that balances agent workloads based on both computation and communication capacities. Recognizing heterogeneous environments where agent computation and communication capabilities fluctuate, ComDML leverages a dynamic, decentralized pairing scheduler. This scheduler pairs agents and assigns workloads based on observed capabilities, ensuring efficient computation and communication in heterogeneous settings.
The scheduler prioritizes pairing the slowest agents first by maintaining a shared list of training times. This list guides agents to pair up at each round, starting with the slowest, to minimize the overall training time. 
This pairing strategy employs lightweight, low-overhead local split model profiling, which quantifies the communication overhead (in terms of intermediate data size) for various pairing options.
To make workload offloading decisions, slower agents consider both local profiling and the communication and computation capacities of faster agents, ensuring optimal workload balancing for the shortest total training time. In this way, the proposed pairing scheduler operates in a decentralized manner with minimal information exchange among agents.

Using standard assumptions in FL \cite{li2019convergence, karimireddy2020scaffold} and local-loss-based training techniques \cite{belilovsky2020decoupled, han2021accelerating}, we show the convergence of ComDML for both convex and non-convex functions. The convergence analysis is novel as it considers multiple split models for each paired agent in a heterogeneous environment. Using ComDML, we conduct training experiments on large models (ResNet-56 and ResNet-110 \cite{he2016deep}) across various numbers of agents using popular datasets including CIFAR-10 \cite{krizhevsky2009learning}, CIFAR-100 \cite{krizhevsky2009learning}, and CINIC-10 \cite{darlow2018cinic}, alongside their non-identical and independent distribution (non-I.I.D.) variants. 
Our extensive experimental results demonstrate that ComDML achieves a remarkable reduction in overall training time by up to 71\% while maintaining model accuracy comparable to state-of-the-art methods. We also evaluate its performance under various privacy measures. These measures include minimizing distance correlation between raw data and intermediate representations, shuffling data patches, and applying differential privacy to model parameters. Our results demonstrate that ComDML can effectively incorporate these privacy techniques with minimal impact on model accuracy.

\section{Background and Related Work}

\subsection{Collaborative Multi-agent Learning}

Federated learning (FL) is a privacy-preserving machine learning technique that allows multiple parties (clients or agents) to collaboratively train a machine learning model without having to share their data \cite{mcmahan2017communication}. This is achieved through frequent communication with a central server for model exchange and updates, facilitating the collaborative learning process \cite{kairouz2021advances}. State-of-the-art deep learning models (e.g., ResNet or AlexNet) have become increasingly large in recent years, which can make the computation and communication cost of FL prohibitive. The computational cost of resource-constrained agents and the communication overhead of FL can become significant challenges, particularly for large models \cite{diao2020heterofl}. 
\cite{wu2022communication} reduces communication overhead by transmitting smaller models, leading to faster training and lower resource consumption. 
Meanwhile, \cite{mohammadabadi2023speed} employs SL to split the global model across agents in different tiers, resulting in faster training times.
In a related context, \cite{chen2021communication} specifically addresses device selection for communication purposes.
In addition to communication optimization, \cite{zhang2021optimizing} studies resource allocation strategies. Leveraging network pruning on client models, \cite{li2024model} improves inference performance, by reducing model size and communication volume.
Notably, these algorithms require a central server and are not designed for decentralized systems. In contrast, ComDML does not require a central coordinator, and it addresses heterogeneity by workload balancing.

\subsection{Server-less and Peer-to-Peer Decentralized Learning}

Existing FL methods (see a comprehensive study of FL  \cite{kairouz2021advances}) have traditionally relied on a central server to manage agent selection, model broadcasting, aggregation, and updating. In these methods, agents are required to repeatedly download and update the global model and send it back to the server. 
However, such processes face limitations when training large models on resource-constrained devices in heterogeneous environments, leading to issues like the straggler problem. Additionally, they are susceptible to vulnerabilities stemming from potential failures of the central server and network bottlenecks.
To address the straggler problem,  \cite{li2019convergence} selects a smaller set of agents for training in each global iteration, but at the cost of increased training rounds. \cite{bonawitz2019towards} deals with stragglers by ignoring the slowest 30\% of agents, while FedProx \cite{li2020federated} assigns different numbers of training rounds to agents. These approaches face the challenge of determining optimal parameters (i.e., percentage of slowest agents and number of local epochs).

To tackle the challenge of central server failures, \cite{roy2019braintorrent} proposes a server-less framework for cross-silo FL, targeting scenarios with a relatively small number of agents with powerful communication and computation capabilities to enable sequential operation. Recent research has focused on fully decentralized approaches, such as the work by \cite{sun2022decentralized}, which aims to achieve fully decentralized learning. 
Gossip learning, as introduced in \cite{ormandi2013gossip}, offers an alternative approach to serverless decentralized learning by enabling devices to exchange model updates with their neighbors. Building on this concept, \cite{hegedHus2019gossip} explores the application of gossip learning as a substitute for FL. GossipFL \cite{tang2023gossipfl} introduces a sparsification algorithm to reduce agent communication to a single peer with a compressed model in a gossip learning setting. Another way to eliminate the central server is the integration of blockchain technology. For instance, \cite{li2020blockchain} proposes a blockchain-based solution with committee consensus, while \cite{ramanan2020baffle} investigates the utilization of blockchain in combination with Smart Contracts. However, it is important to note that these methods do not explicitly consider agent heterogeneity and require agents to train the entire global model. The learning performance may degrade a lot due to the straggler problem, particularly in scenarios where agents have varying computation/communication capabilities or limited resources. In this paper, we address these issues by developing a decentralized workload balancing algorithm that does not rely on a central server and can effectively pair agents in dynamic heterogeneous environments.

\section{Workload Balancing for Decentralized Multi-Agent Learning}

\begin{figure}[t]
  \centering
   \includegraphics[width=1.00\linewidth]{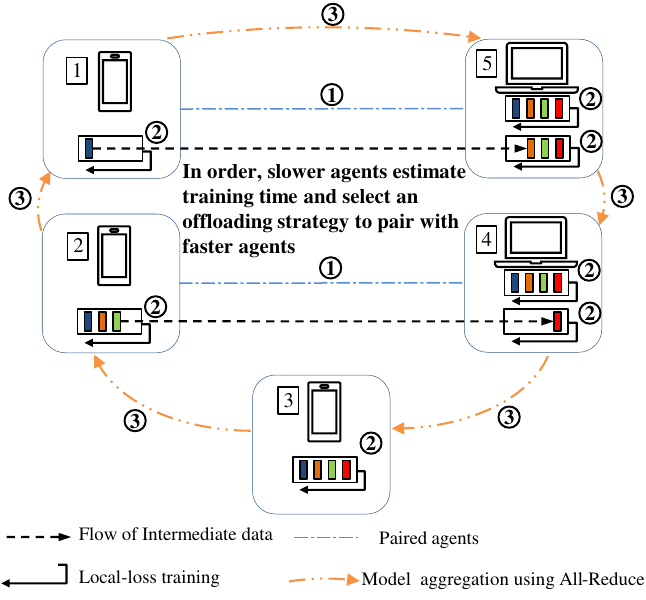}

   \caption{Overview of the training process for ComDML. In each round, ComDML first pairs agents \textcircled{1}. The paired agents then carry out the local model update \textcircled{2}. After all agents complete the model update, ComDML uses a decentralized model aggregation to obtain the global model \textcircled{3}. In the above example, agents \framebox{1} and \framebox{2} offload work to faster agents \framebox{5} and \framebox{4}, respectively, while agent \framebox{3} trains independently.}  
   
   \label{fig:pairing_scheduler}
\end{figure}

This work focuses on DML systems as introduced by \cite{lalitha2019peer}, where multiple agents collaborate on a learning task within a networked environment, absent a central coordinating server. 
Beyond applications in healthcare, mobile services, and vehicle networks, DML is fueling advancements in emerging areas like swarm robotics, smart cities, and the metaverse \cite{beltran2023decentralized}.
Within a DML system, multiple agents collaboratively train a global model using their local datasets and then synchronize with other agents to update the model. In practice, agents usually have heterogeneous computation resources as well as heterogeneous data, thus the training workload across agents can be highly different, which leads to very different training times. Therefore, faster agents may have to wait for slower agents (stragglers) for a long time to synchronize the model update. Such a bottleneck in synchronization may not only significantly increase the overall training time, but also result in a waste of the spare computation resources of faster agents.

To address this challenge, one promising solution is to balance the training workload based on the computation and communication resources by offloading the workload from slower agents to faster agents, to address the straggler problem. Fig. \ref{fig:pairing_scheduler} illustrates this workflow. It is worth noting that the training workload of each agent is highly correlated with the local dataset size of each agent. When offloading the workload from slower agents to faster agents, faster agents still need to frequently communicate with slower agents to exchange intermediate training information, and the amount of communication depends on how much workload is offloaded. Due to the heterogeneous communication capacity between agents, the communication overhead may offset the benefits of workload balancing. Therefore, optimal workload balancing requires jointly considering the heterogeneous communication and computation resources of agents. In this paper, we aim to develop a communication-efficient training workload balancing approach for collaborative multi-agent learning in a decentralized system.

\subsection{Decentralized Multi-agent Learning}

Consider $K$ agents in a decentralized system (see Fig. \ref{fig:pairing_scheduler}), where $\left\{\left(\boldsymbol{x}_n, y_n\right)\right\}_{n=1}^{N_{i}}$ denotes the dataset of agent $i$. Here, $\boldsymbol{x}_n$ represents the $n$th training sample, $y_n$ is the associated label, and $N_{i}$ is the number of samples in agent $i$'s dataset.

For convenience, define $f_{i}(\bm{w}) =\frac{1}{N_{i}} \sum_{n=1}^{N_{i}} \ell\left((\boldsymbol{x}_n, y_n\right); \bm{w})$, with $\bm{w}$ being the model parameters. The DML problem can be formulated as a decentralized optimization problem:
{  \begin{align} 
\label{eq:glob_obj}
\min _{\bm{w}} f(\bm{w}) &\stackrel{\text { def }}{=} \min _{\bm{w}} \sum_{i=1}^K \frac{N_{i}}{N} \cdot f_{i}(\bm{w})
\end{align}}

\noindent where   $N=\sum_{i=1}^K N_{i}$. $f(\bm{w})$ denotes the global objective function, and $f_{i}(\bm{w})$ represents the $i$th agent's local objective function, which measures the individual loss over its heterogeneous dataset using a  loss function $\ell$. Each agent possesses its local data and collaborates with other agents to find the optimal $\bm{w}$ that minimizes the global objective (\ref{eq:glob_obj}).

Federated optimization techniques (e.g., \cite{mcmahan2017communication, li2020federated}) have been proposed to solve the problem  (\ref{eq:glob_obj}). 
However, these FL methods encounter challenges in training models on resource-constrained devices, especially large models in heterogeneous environments. In particular, faster devices have to wait for straggler devices that take longer time to complete tasks, which would slow down the overall training process and waste the spare computation resources of faster devices.
Furthermore, the central server, both a bottleneck and a prime target for potential attacks \cite{lyu2020threats}, poses a significant risk of disrupting the training process through failures or downtime.
To address this challenge, we consider a DML system without a central server and enable workload balancing among agents.

\subsection{Workload Balancing via Local-loss based Split Training}\label{sec:local-loss}

To achieve efficient workload balancing for DML, we employ local-loss-based split training. 
Specifically, the model $\bm{w}=(\bm{w}_i^{a_s^m},\bm{w}_i^{a_f^m})$ is split into two parts: a slow agent-side model $\bm{w}_i^{a_s^m}$ and a fast agent-side model $\bm{w}_i^{a_f^m}$. This split model allows slower agent $i$ to train only the slow agent-side model $\bm{w}_i^{a_s^m}$ and an auxiliary network $\bm{w}_i^{{aux}^m}$. The auxiliary network consists of additional layers connected to the slow agent-side model and is used to compute the local loss on the slow agent-side model. 
By incorporating the auxiliary network, we enable parallel model updates for each agent \cite{han2021accelerating}, avoiding the significant synchronization and communication overhead associated with split learning \cite{vepakomma2018split}, which can significantly slow down the training process. In this paper, we adopt the approach of employing a few fully connected layers for the auxiliary network, following the approach in \cite{han2021accelerating,belilovsky2020decoupled}.
Consider $M$ split models to determine how a slower agent splits the model for offloading to a faster agent. By offloading a portion of the model to the faster agent, the workload on the slower agent reduces, aiming to achieve equal training time for the paired agents.

Given the model $\bm{w}$, we define $f^{a_s^m}_i(\bm{w}_i^{a_s^m},\bm{w}_i^{{aux}^m})$ as the loss function of the slow agent-side and $f^{a_f^m}_i(\bm{w}_i^{a_f^m},\bm{w}_i^{a_s^m})$ as the corresponding fast agent-side loss function for the paired agents, where $m$ denotes how the slower agent splits the model to be offloaded to the faster agent. The goal of the slower agent $i$ is to find $\bm{w}_i^{a_s^{m^{\star}}}$ and $\bm{w}_i^{{aux}^m}$ that minimize the loss function on the slow agent-side for each paired agent.

{  \begin{align} \label{eq:slow_agent_obj}
 & \min _{\bm{w}_i^{a_s^m},\bm{w}_i^{{aux}^m}} f^{a_s^m}_i(\bm{w}_i^{a_s^m},\bm{w}_i^{{aux}^m}) = \nonumber \\ & \min _{\bm{w}_i^{a_s^m},\bm{w}_i^{{aux}^m}} \frac{1}{N_{i}} \sum_{n = 1}^{N_{i}} \ell\left((\boldsymbol{x}_n, y_n); (\bm{w}_i^{a_s^m},\bm{w}_i^{{aux}^m})\right)
\end{align}}

The faster agent $j$ shares its computation resources to simultaneously optimize $\bm{w}_j$ by minimizing local objective function and find $\bm{w}_i^{a_f^m\star}$ that minimizes \eqref{eq:fast_agent_obj}:
{  \begin{align} \label{eq:fast_agent_obj}
 & \min _{\bm{w}_i^{a_f^m}} f^{a_f^m}_i(\bm{w}_i^{a_f^m},\bm{w}_i^{a_s^m{\star}}) = \nonumber \\ & \min _{\bm{w}_i^{a_f^m}} \frac{1}{N_{i}} \sum_{n = 1}^{N_{i}} \ell\left((\boldsymbol{z}_n, y_n); (\bm{w}_i^{a_f^m},\bm{w}_i^{a_s^m{\star}})\right)
\end{align}}

\noindent where $\boldsymbol{z}_n$ denotes the intermediate output of the slow agent-side model $\bm{w}_i^{a_s^m{\star}}$ given the input $\boldsymbol{x}_n$.
This process continues until all agents have completed their training tasks in a training round. At the end of each round, the models are aggregated using the AllReduce method \cite{goyal2017accurate} (see Sec. \ref{sec:workflow}).

\subsection{Optimization for Workload Balancing}

In a decentralized system with multiple agents, the objective of workload balancing is to minimize the overall training time.
To this end, two key questions need to be addressed: 1) how to pair a faster agent with a slower agent for workload balancing, and 2) how much workload to offload when two agents are paired. Using the proposed local-loss-based split training for workload balancing, we need to jointly optimize the communication and computation time when addressing these two key questions.

Specifically, let $\gamma_{ij}\in\{0,1\}$ denote the workload balancing decision, and $\gamma_{ij}=1$ means offloading agent $i$'s workload $\bm{w}_i^{a_f^m}$ to agent $j$ by using split model $m$. 
The overall training time $\tau_i$ for agent $i$ in each round can be presented as:

 {
\begin{equation} \label{eq:time}
\tau_i=\left\{
	\begin{array}{*{20}c}
        \tau_i(\bm{w})+ \sum_j \gamma_{ji}\left[\tau_{ji}(\bm{w}_i^{a_f^m}) + \tau_i(\bm{w}_j^{a_f^m})\right], \\ \text{if}~\sum_j \gamma_{ij}=0 \text{ \small (Agent $i$ does not offload workload)}\\
        \tau_i(\bm{w}_i^{a_s^m}), \\ \text{if}~\sum_j \gamma_{ij}=1 \text{ \small (Agent $i$  offloads workload to one agent)}	,	\end{array} \right.
\end{equation}
}

\noindent where $\tau_i(\bm{w})$ denotes agent $i$'s computation time of learning the model $\bm{w}$. $\tau_{ij}(\bm{w}_i^{a_f^m})$ denotes the communication time when agent $i$ offloads its workload to agent $j$, which depends on the speed of communication link between agents $i$ and $j$, as well as the amounts of intermediate data based on $\bm{w}_i^{a_s^m}$. 
In (\ref{eq:time}), if agent $i$ does not offload workload, the overall training time consists of $\tau_i(\bm{w})$ and the computation and communication time for processing the workload of slower agents $\sum_j \gamma_{ji}\left[\tau_i(\bm{w}_j^{a_f^m}) +  \tau_{ji}(\bm{w}_i^{a_f^m}) \right]$ if any; if agent $i$ offloads workload, the overall training time consists of the computation time of learning a slow agent-side model $\tau_i(\bm{w}_i^{a_s^m})$. and the corresponding communication time $\sum_j \gamma_{ij}\tau_{ij}(\bm{w}_i^{a_f^m})$ with a faster agent.

The problem of joint optimization of communication and computation for workload balancing can be formulated as  minimizing the training time of the slowest agent (i.e., straggler):

{
\begin{equation} \label{eq:workload}
\begin{array}
		[c]{lll}%
\min\limits_{\{\gamma_{ij}\},\{\bm{w}_i^{a_f^m}\}} \max\limits_i \tau_i,  \ \  \ \
\text{s.t.} ~\gamma_{ij}\in\{0,1\} ~~\forall i,j.
\end{array}
\end{equation}}

\noindent In (\ref{eq:workload}), we need to jointly optimize the workload balancing decisions $\{\gamma_{ij}\}$ and the offloaded workload (i.e., how to split the model $\bm{w}=(\bm{w}_i^{a_f^m},\bm{w}_i^{a_s^m})$ when $\gamma_{ij}=1$). Note that problem (\ref{eq:workload}) is an integer programming problem, which is challenging to solve in a decentralized learning system without any centralized scheduler.

\section{Decentralized Workload Balancing}

\subsection{Agent Pairing} \label{sec:scheduler}

To effectively address the optimization problem in (\ref{eq:workload}), we require a pairing strategy that dynamically pairs agents based on their computation and communication capacities in each training round. This dynamic approach is crucial due to the inherent variability of agent capabilities within heterogeneous environments. Static pairing assignments can lead to significant straggler problems, as agents with limited resources may be paired together, or slow agents may remain unpaired. Such pairings can inadvertently increase the overall training time.
To mitigate these issues, we propose a dynamic decentralized pairing scheduler. This scheduler dynamically pairs agents in a decentralized manner to minimize overall training time based on agents' computation and communication capacities.

As agents train their models in parallel, the training time in each round is determined by the slowest agent, denoted as $\max\limits_i \tau_i$. This training time for the slowest agent becomes a critical factor, as other agents must wait for it before model aggregation can commence. To minimize the overall training time, our objective is to minimize the maximum training time among all agents, as expressed in (\ref{eq:workload}). 
Leveraging a list of individual training times, updated and maintained by each agent, agents dynamically pair themselves each round, prioritizing the slowest agent first, to minimize the overall training time for each pair. In this approach, agents are paired or trained independently based on their resources and available neighbors. 
To ensure optimal pairings, the process meticulously considers the communication link speed, processing speed, and dataset size of a faster agent. This decentralized approach empowers each agent to independently implement the pairing scheduler, fostering scalability and resilience without reliance on a central coordinator. Information exchange is minimized, as agents only need to share their processing speeds and local dataset sizes with their neighbors. Individual training times are calculated based on these shared metrics, and network speeds can be directly observed.

\subsection{Training Workflow} \label{sec:workflow}

The training process of ComDML in each round is described in the following steps, which are detailed in Algorithm \ref{alg:pairing_scheduler} and illustrated in Fig. \ref{fig:pairing_scheduler}. This algorithm achieves remarkable resource optimization with minimal overhead, enabling agents to fully leverage spare resources for significant performance gains.

To facilitate the decentralized agent pairing, each agent locally conducts split model profiling prior to the training process. The split model profiling calculates the relative training time (i.e., the training time compared to the case where the model is not split) and intermediate data size for each split model $m$. Specifically, for $M$ different split models, each agent calculates the relative training time of the slow agent-side $T^{a_s^m}$, the fast agent-side $T^{a_f^m}$ and the intermediate data size $\nu^m$ of each split model $m$ using a batch with the same size. The profiling helps each agent to estimate the overall training time of each split model based on the actual size of the dataset when pairing.

\textbf{\textcircled{1} Agent pairing.} In each round, slower agents initially pair up in order of their estimated training times. The agent pairing algorithm ensures that each pair minimizes their training time and completes their tasks within a similar time.
Specifically, all available agents broadcast their processing speed $p_j$ and individual training time $\tau_j$ (i.e., time required to complete its task without workload offloading) to all connected agents in their network. 
Through a greedy algorithm (function \textbf{Pairing($\cdot$)}), agents are paired in order of their individual training times. Starting with the agent with the longest training time, each agent selects a faster agent that can significantly reduce its training time by offloading part of its workload. This pair then informs the next agent in the list to pair up. This ensures that paired agents complete their training in similar time.
Agent $i$ estimates the overall training time if it offloads its workload to agent $j$, using the \textbf{AgentTrainingTime($\cdot$)} function. 
This function factors in the processing speeds of both agents $p_i$ and $p_j$, agent $j$'s estimated individual training time $\hat{\tau}_j$, the network speed $c_{ij}$, and and the data transfer size during offloading. To estimate the training time for split model $m$, agent $i$ utilizes the split model profile to convert $p_i$ and $p_j$ into the processing speeds on the slow agent-side $p_i^m$ and the fast agent-side $p_j^m$ of model $m$, respectively. 
Let $\Tilde{N}_i$ denote the number of data batches of agent $i$. Agent $i$ estimates the time for different split model $m$ as follows: $\hat{\tau}_{ij}^m = {\max\left(\frac{\Tilde{N}_i}{p_i^m},  \hat{\tau}_{j}+ \frac{\Tilde{N}_i\nu^m}{c_{ij}} + \frac{\Tilde{N}_i}{p_j^m}\right)}$, where $\hat{\tau}_{ij}^m$ represents the estimated training time for agent $i$ when using split model $m$. This process is implemented in the \textbf{Pairing($\cdot$)} function in Algorithm \ref{alg:pairing_scheduler}.

\textbf{\textcircled{2} Local model update.} Then, each pair of agents collaboratively perform the slower agent's task via local-loss-based split training (see Sec. \ref{sec:local-loss}). Simultaneously, each faster agent also performs the model training using its local dataset.

\textbf{\textcircled{3} Model aggregation and update global model.}
At the end of each round $r$, all agents participate in the AllReduce operation \cite{goyal2017accurate} to synchronize their models and obtain the average of all agents' models. Following the aggregation process (i.e., the \textbf{ModelAggregation($\cdot$)} function), all agents have the updated model parameters that represent the average of all $K$ agents. 
The AllReduce mechanism facilitates a key decentralized aspect of the aggregation process. It allows for the sharing and averaging of updated model weights among agents without the need for a centralized coordinator.
Two well-known AllReduce algorithms suitable for bandwidth-limited scenarios are the recursive halving and doubling algorithm 
\cite{thakur2005optimization} and the ring algorithm \cite{goyal2017accurate}. In both algorithms, each agent sends and receives $2\frac{K-1}{K}b$ bytes of data, where $b$ represents the model size in bytes. The halving/doubling algorithm consists of $2\log_2(K)$ communication steps, while the ring algorithm involves $2(K-1)$ steps. Given that we are dealing with a large number of agents, we opt for the halving and doubling algorithms for the AllReduce operation. Other existing aggregation techniques (e.g., quantized gradients \cite{hubara2017quantized}) can also be integrated into the proposed training process to further reduce communication overhead.

\begin{algorithm}[t]
\caption{\textbf{ComDML.}  
}\label{alg:pairing_scheduler}
\begin{algorithmic}[1]

\Statex \textbf{Initialize:} $R$ denotes the total global rounds, $T^{a_s^m}$ and $T^{a_f^m}$ denote the relative training time for the slow and fast agent-side sides, respectively, corresponding to model split $m$ with an intermediate data size of $\nu^m$, $\mathcal{A}$ is the list of descending order of agents by their task completion times, 
$\left[\hat{\tau}_{ij}^{m}\right]$ denotes the list of estimated training time. 

\Statex \textbf{Main()}
\For{$r=0$ to $R-1$}
    \State Agents broadcast $p_j$ and $\hat{\tau_{j}}$ to all connected agents
    \For{agent $i$ in order $\mathcal{A}$}
        \If{agent $i$ is not paired}
            ${j^\star} \leftarrow \textbf{Pairing($i$)}$
        \EndIf
    \EndFor
    \State \textbf{Local model update across agents in parallel}

    \State \textbf{ModelAggregation()} \Comment{Decentralized aggregation}
\EndFor

\Statex \textbf{Pairing($i$)} \Comment{Run on agent $i$}
 
\For{all unpaired $j$ that are connected to $i$}
    \Statex // Estimate the training time of $i$ if it offloads to $j$
    \State $\hat{\tau}_{ij} \gets \textbf{AgentTrainingTime}(p_j, \hat{\tau_{j}})$

\EndFor

\State $j^\star \leftarrow \arg \min\limits_{j} \left[\hat{\tau}_{ij}\right]$
\State \textbf{Return} $\gamma_{ij^\star}$

\Statex \textbf{AgentTrainingTime($p_j$, $\hat{\tau}_{j}$)} \Comment{Run on agent $i$}
\For{all split layer $m$}

    \State $p_i^m \gets  \frac{p_i}{T^{a_s^m}}$
    \State $p_j^m \gets  \frac{p_j}{T^{a_f^m}}$
    \State $\hat{\tau}_{ij}^{m} \gets {\max\left(\frac{\Tilde{N}_i}{p_i^m},  \hat{\tau}_{j}+ \frac{\Tilde{N}_i\nu^m}{c_{ij}} + \frac{\Tilde{N}_i}{p_j^m}\right)}$
\EndFor

\State $\hat{\tau}_{ij} \gets \min \limits_{m} \left[\hat{\tau}_{ij}^{m}\right]$

\State $m^\star \gets \arg \min \limits_m \left[\hat{\tau}_{ij}^{m}\right]$ 

\State \textbf{Return} {$\hat{\tau}_{ij}$}

\end{algorithmic}
\end{algorithm}

\subsection{Privacy Protection}
\label{sec:privacy}

While ComDML excels in reducing training time, it addresses privacy concerns arising from model intermediate data exchange.
To mitigate model parameter attacks that aim to replicate models using dummy data inputs \cite{shen2023ringsfl}, ComDML restricts agents' access to external datasets, query services, and dummy data itself, effectively shielding sensitive information from potential adversaries.
Furthermore, ComDML's model split architecture, solidified by AllReduce aggregation, inherently counters model inversion attacks by compartmentalizing model updates between slow and fast agents. This architectural design restricts model visibility, aligning with research that suggests such attacks often require full model access to succeed \cite{yin2021see}.

While ComDML's architecture inherently limits privacy leakage, it acknowledges potential vulnerabilities to strong eavesdropping attacks. To address this, it offers a versatile framework that seamlessly integrates with diverse privacy-preserving techniques: 
\begin{inparaenum}[\itshape i\upshape)]
    \item \textbf{Fast agents privacy}: Inherently protected through unidirectional communication (i.e., from slow agents to fast agents), fast agent updates remain confidential. For further privacy guarantees during model aggregation, techniques like differential privacy \cite{abadi2016deep} and cryptography \cite{sami2023secure} can be integrated.
    \item \textbf{Slow agents privacy}: ComDML prioritizes slow agent privacy with a customizable toolkit. Techniques like PixelDP noise layer \cite{lecuyer2019certified}, patch shuffling \cite{yao2022privacy}, distance correlation \cite{vepakomma2020nopeek}, and  SplitGuard \cite{erdogan2022splitguard}  directly shield intermediate data, while differential privacy or cryptography secure model aggregation. This flexibility empowers slow agents to tailor their protection, balancing privacy and performance.

\end{inparaenum}

\subsection{Convergence Analysis}
\label{sec:convergence}

We establish the convergence of slow and fast agent-side models, considering both convex and non-convex loss functions under standard assumptions. This is achieved through local-loss-based training adapted from \cite{belilovsky2020decoupled}, where input distributions for fast agents dynamically evolve based on the convergence of their slow counterparts. We define $A^{m,r}$ and $\mathcal{A}^{m,r}$ as the number and the set of agents with split model $m$ at round $r$, respectively. The output of the slow agent-side model, $\boldsymbol{z}_n^{a_s^m,r}$, follows the density function $d^{a_s^m,{r}}$, where the converged density of the slow agent-side is represented as $d^{a_s^m,{\star}}$. We define $c^{a_s^m,{r}} \triangleq \int\left|d^{a_s^m,{r}}(\boldsymbol{z})-d^{a_s^m,{\star}}(\boldsymbol{z})\right| d\boldsymbol{z}$ as the distance between the density function of the output of the slow agent-side model and its converged state. In the following, we introduce the standard assumptions used in the analysis.

\begin{as}[L-smoothness] \label{as:l_smooth}
The loss function $f$ is differentiable and $L$-smooth, i.e.,  $\left\|\nabla f_i(\bm{w})-\nabla f_i(\boldsymbol{v})\right\| \leq L\|\bm{w}-\boldsymbol{v}\|$, $\forall$ $f_i$, $\bm{w}$, $\boldsymbol{v}$.
\end{as}

\begin{as}[$\mu$-convex] \label{as:mu_convex}
$f_i$ is $\mu$-convex for $\mu \geq 0$ and satisfies:
$
f_i(\bm{w}) + ( \boldsymbol{v}-\bm{w})^T \nabla f_i(\bm{w}) +\frac{\mu}{2}\|\boldsymbol{v}-\bm{w}\|^2 \leq f_i(\boldsymbol{v}), \forall f_i, \bm{w}, \boldsymbol{v} .
$

\end{as}

\begin{as}[Bounded gradients] \label{as:bound_gradient} 
Expected squared norms of gradients have upper bounds:
 $\mathbb{E}\left\|\nabla f_i(\bm{w})\right\|^2 \leq G_1^2, \forall f_i, \bm{w}$. 
\end{as}

\begin{as}[Bounded variance] \label{as:bound_variance}
The variance of stochastic gradients in each agent is bounded: $\mathbb{E}[\|\nabla f_i(\zeta_i^r, \bm{w}) - \nabla f_i(\bm{w})\|^2] \leq \sigma^2, \forall f, \bm{w}$, where $\zeta_i^r$ sampled from $k$-th agent local dataset.

\end{as}

\begin{as}[Bounded gradient dissimilarity] \label{as:gradient_diss}
For both slow and fast agent-sides and all split models, there are constants $G_2 \geq 0$; $B \geq 1$ such that
$
\frac{1}{K} \sum_{i=1}^K\left\|\nabla f_i(\bm{w})\right\|^2 \leq G_2^2+B^2\|\nabla f(\bm{w})\|^2, \forall \bm{w}.
$
\end{as}

If $\left\{f_i\right\}$ are convex, we can relax the assumption to
$
\frac{1}{K} \sum_{i=1}^K\left\|\nabla f_i(\bm{w})\right\|^2 \leq G_2^2+2 L B^2\left(f(\bm{w})-f^{\star}\right), \forall \bm{w} .
$

\begin{as}[Bounded distance]\label{as:conv_pre_layer}
The time-varying parameter satisfies $ \sum_r c^{a_s^m,{r}} <\infty$,  $\forall m$. 
\end{as}

Assumptions 1 to 6 have been widely employed in the literature for convergence analysis of machine learning (see \cite{li2019convergence,belilovsky2020decoupled,karimireddy2020scaffold} and the references therein). Under these standard assumptions, we establish the convergence properties of ComDML. The proof of Theorem \ref{thm:slow_side} is given in the Appendix.

\begin{thm}[Convergence of ComDML]\label{thm:slow_side}
Suppose that $f^{a_s^m}$ and $f^{a_f^m}$ satisfy Assumptions \ref{as:l_smooth}, \ref{as:bound_gradient}, \ref{as:bound_variance}, \ref{as:gradient_diss}, and \ref{as:conv_pre_layer}. 
The convergence properties of ComDML for both convex and non-convex functions are summarized as follows:

\begin{itemize}[leftmargin=*]

\item \textbf{Convex:} If both $f^{a_s^m}$ and $f^{a_f^m}$ are $\mu$-convex with $\mu > 0$, $\eta \leq \frac{1}{8L(1+B^2)}$ and $R \geq \frac{4L(1+B^2)}{\mu}$, then the slow agent-side model converges at the rate of $\mathcal{O}\left(\frac{H_1^2 }{{R A^m}}+ D^2\exp\left(-{R}\right)\right)$ and the fast agent-side model converges at the rate of $\mathcal{O}\left(\frac{H_2 \sqrt{F^{a_f^m{0}}} }{\sqrt{R A^m}} + \frac{C_1 + F^{a_f^m{0}}}{R}  \right)$.

\item \textbf{Non-convex:} If both $f^{a_s^m}$ and $f^{a_f^m}$ are non-convex with $\eta \leq \frac{1}{8L(1+B^2)}$, then the slow agent-side model converges at the rate of $\mathcal{O}\left(\frac{H_1 \sqrt{F^{a_s^m{0}}}}{\sqrt{R {A^m}}}+\frac{F^{a_s^m{0}}}{R}\right)$ and the fast agent-side model converges at the rate of $\mathcal{O} \left( \frac{H_2\sqrt{{F^{a_f^m{0}}} }}{\sqrt{R{A^m}}}  + \frac{C_2 + {F^{a_f^m{0}}}}{ R}\right)$.

\end{itemize}
where $H_1$, $H_2$, $D$, $F^{a_s^m}$, $F^{a_f^m}$, and $A^{m}$ are constants whose definitions are provided in the Appendix for reference. 
Demonstrated slow agent convergence propagates to fast agents through $C_1$ and $C_2$, both exhibiting convergence.

\end{thm}

We demonstrate that under standard assumptions, ComDML exhibits convergence for both convex and non-convex functions as the number of training rounds $R$ increases. This convergence behavior holds for both slow and fast agent-side models, albeit with potentially different convergence rates for fast and slow agent sides. It's crucial to note that ComDML's reliance on local-loss-based split training renders the convergence of the fast agent-side model contingent upon the convergence of the slow agent-side model. This dependence is explicitly characterized by constants $C_1$ and $C_2$ within the analysis.

\section{Experimental Evaluation}
\label{sec:experiments}

\subsection{Experimental Setup}
\noindent\textbf{Dataset.} We conduct image classification experiments using three publicly available datasets: CIFAR-10 \cite{krizhevsky2009learning}, CIFAR-100 \cite{krizhevsky2009learning}, and CINIC-10 \cite{darlow2018cinic}.  
We also consider label distribution skew (i.e., how the distribution of labels varies across agents) to generate non-I.I.D. variants of these datasets. To maintain fairness, we used a fixed Dirichlet distribution (concentration parameter = 0.5) for the non-I.I.D. datasets. Global model performance is assessed using test images after each round.

\noindent\textbf{Baselines.}
To the best of our knowledge, this study pioneers the introduction of workload balancing in server-less DML. Although FL methods like FedProx \cite{li2020federated} aim to enhance performance in heterogeneous environments, they rely on a central server, incompatible with our focus on serverless decentralized machine learning. Therefore, we primarily compare ComDML with decentralized baselines: BrainTorrent \cite{roy2019braintorrent}, Gossip Learning \cite{hegedHus2019gossip}, and decentralized AllReduce \cite{goyal2017accurate}. 
FedAvg \cite{mcmahan2017communication}, though server-dependent, is included as a baseline for comparing workload balancing.
BrainTorrent is a peer-to-peer framework where agents take turns acting as the server and updating the global model. Gossip Learning \cite{hegedHus2019gossip} incorporates model averaging, enabling each agent to update its model based on information received from neighboring agents. In decentralized learning utilizing AllReduce aggregation, agents update their models independently and then employ AllReduce to aggregate them, eliminating the need for a central server.

\noindent\textbf{Implementation.}
We conducted the experiment using Python 3.11.3 and the PyTorch library version 1.13.1. The source code is publicly available online at ComDML's GitHub repository \cite{code}. ComDML and baseline models were deployed on a server with the following specifications: dual-socket Intel(R) Xeon(R) CPU E5-2630 v4 @ 2.20GHz, four NVIDIA GeForce GTX 1080 Ti GPUs, and 64 GB of memory.
To replicate the heterogeneity of real-world systems, we designed a heterogeneous simulation environment where agents possess distinct computational and communication capabilities. Each agent is equipped with a simulated CPU and communication resources, mirroring varied computation and communication times. This setup effectively captures the complexities of real-world distributed systems, where agents often differ in their processing power and network connectivity.
We simulated agents with CPU profiles spanning 4, 2, 1, 0.5, and 0.2 CPUs. Communication profiles were varied with 0, 10, 20, 50, and 100 Mbps links, with 0 representing disconnected agents. At the beginning of training, each agent was assigned one profile representing its initial computation and communication resources. These resource profiles could dynamically shift throughout the training process, mimicking real-world variations in agents' resources.
In all experiments, we consider simulated communication overhead for training time, including intermediate data transfer and model size.
BrainTorrent operates through independent model updates followed by aggregation by a randomly selected agent. Gossip learning implementation aligns with \cite{hegedHus2019gossip}. 
For AllReduce experiments, independent model training is followed by aggregation via AllReduce.

\noindent\textbf{Model Architecture.} 
ComDML can effectively support various models, from Multilayer Perceptrons (MLPs) and Convolutional Neural Networks (CNNs) to large language models (LLMs) like BERT.
In our experiments, we used two prominent CNN models: ResNet-56 and ResNet-110 \cite{he2016deep}, which have demonstrated good accuracy on the experimented datasets. To balance computational demands across agents, we partition the global model at varying split layers $m$. Slow agents locally update the model up to their designated layer $m$, while a fast agent handles the remaining layers. To facilitate local loss training within slow agents, we introduce an auxiliary network comprising a fully connected layer and an average pooling layer. The input dimension of the f.c. layer is adjusted to match the output of each slow agent-side model.

\noindent\textbf{Hyper-parameters.} 
We used the Stochastic Gradient Descent (SGD) optimizer with a momentum of 0.9 for all datasets. The initial learning rate (i.e., $\eta_0$) was set to 0.001 for all datasets. Upon the accuracy reached a plateau, the learning rate was reduced by a factor of 0.2 when there were 10 agents. For scenarios with 20, 50, and 100 agents, we implemented a reduction factor of 0.5. The local batch size for each agent was set to 100, and the local epoch was consistently set to one for all experiments.

\begin{center}
\begin{table}[t]
    \caption{Performance of 2-agent decentralized training with varying layer offloading. The evaluation compares the fast agent training time, communication time, combined idle time of agents, and total training time of the process (all in seconds) required to achieve 90\% accuracy on CIFAR-10 using the ResNet-56.}
    \label{tab:offload}
    \centering
    \setlength{\tabcolsep}{3pt}
    \begin{tabular}{c | c c c c | c c c c  } 
        \toprule
        \multirow{2}{*}{\shortstack[c]{Layers\\Offloaded}} & \multicolumn{4}{c|}{Times in the 1st Setting (s)} & \multicolumn{4}{c}{Times in the 2nd Setting (s)} \\
        & \multicolumn{1}{c}{Train} & \multicolumn{1}{c}{Comm.} & \multicolumn{1}{c}{Idle} &\multicolumn{1}{c|}{Total} & \multicolumn{1}{c}{Train} & \multicolumn{1}{c}{Comm.} & \multicolumn{1}{c}{Idle} &\multicolumn{1}{c}{Total} \\
        \midrule
        0  & 5573 & 34 & 14489 & 20096 &  5578 & 17 & 3560 &  9165    \\
        1  & 5781 & 655 & 14472 & 20909 & 5856 & 327 & 2966 & 9150   \\
        10 & 6740 & 3532 & 4787 & 15059 & 6364 & 1766 & 350 & 8481   \\
        19 & 7625 & 3544 & 1682 & 12851 & 6547 & 1772 & \textbf{137} & \textbf{8456}  \\
        28 & 7906 & 1261 & 2049 & 11217 & 7859 & 630 & 3368 & 8490   \\
        37 & 8003 & 1265 & \textbf{84} & \textbf{9352} & 8275 & 632 & 7318 & 8908   \\
        46 & 8939 & 611 & 5042 & 9551 & 9334 & 305 & 8351 & 9640  \\
        55 & 10343 & 640 & 9833 & 10983 & 10101 & 320 & 9964 & 10421   \\
        \bottomrule
    \end{tabular}
\end{table}
\end{center}

\subsection{Experimental Results} \label{sec.results}
\subsubsection{Impact of heterogeneity on workload offloading decisions}
Table \ref{tab:offload} compares the performance of local-loss-based split training between two agents, using various portions of the model offloaded from a slower agent to a faster agent in two settings: 1) one agent with 2 CPUs and one agent with 0.25 CPU, with a communication speed of 50 Mbps, and 2)  one agent with 2 CPUs and one agent with 1 CPU, with a communication speed of 100 Mbps.
We evaluate training time for the CIFAR-10 dataset's classification task, targeting 90\% accuracy. Table \ref{tab:offload} presents the fast agent's training time, communication time, the combined idle time of both agents and the overall training time under different workload offloading decisions. The results reveal the significant impact of heterogeneous computation and communication resources on the optimal workload offloading and the total training time.
In ComDML, offloading 0 layers signifies performing the training task independently without assistance. 

Offloading the training workload from the slower agent to the faster agent can effectively reduce the total training time, as demonstrated in Table \ref{tab:offload}. 
For instance, in the 1st setting, offloading 37 layers of the model to the faster agent resulted in a significant 53\% decrease in overall training time compared to the scenario without workload offloading.
Note that as an agent offloads more layers, the model size on its side decreases, reducing the computational workload. However, this offloading of more layers may increase the data transmitted  (i.e., the size of the intermediate data and partial model). 
As shown in Table \ref{tab:offload}, the optimal number of layers to offload is non-trivial, as it depends on various factors, such as the communication link speed between agents, the computation power of each agent, and the workload offloading strategy.
Hence, dynamically pairing agents with suitable offloading strategies during the training process is of paramount importance to achieve substantial reductions in total training time.

\begin{table}[t]
    \caption{Comparison of total training time to baselines with 10 agents on different datasets. The number indicates the training time (in seconds) required to reach the target accuracy. The target accuracies are as follows: CIFAR-10 I.I.D. 90\%, CIFAR-10 non-I.I.D. 85\%, CIFAR-100 I.I.D. 65\%, CIFAR-100 non-I.I.D. 60\%, CINIC-10 I.I.D. 75\%, and CINIC-10 non-I.I.D. 65\%.}
        \label{tab:comparison}
    \centering
    \setlength{\tabcolsep}{3pt}  
    \begin{tabular}{c | c c c c c c c}
        \toprule
        \multirow{2}{*}{Method} & \multicolumn{2}{c}{CIFAR-10} & \multicolumn{2}{c}{CIFAR-100} & \multicolumn{2}{c}{CINIC-10} \\
        &  {I.I.D.} &  {non-I.I.D.} & {I.I.D.} &  {non-I.I.D.} &  {I.I.D.} &  {non-I.I.D.} \\
        \midrule
         {ComDML} & \textbf{7211} & \textbf{4177} & \textbf{5589} & \textbf{8104} & \textbf{10229} & \textbf{17208} \\
         {Gossip Learning} & 20337 & 15269 & 15262 & 28621 & 24636 & 56325 \\
         {BrainTorrent} & 24639 & 14323 & 18046 & 25867 & 31992 & 51144 \\
         {AllReduce} & 25153 & 13859 & 18462 & 26623 & 32652 & 53265 \\
         {FedAvg} & 24174 & 13095 & 17630 & 25113 & 30601 & 49624 \\
        \bottomrule
    \end{tabular}
\end{table}

\subsubsection{Training time improvement against baselines}
In Table \ref{tab:comparison}, we compare the training time of ComDML with baselines by training a ResNet-56 with 10 agents on heterogeneous agents with diverse computation and communication capacities. We created these heterogenous agents by randomly assigning 20\% of the agents to each CPU and communication speed profile combination. All agents participated in the entire training process. To better simulate a dynamic environment, we randomly changed the profile of 20\% of the agents after 100 rounds.
 
The corresponding training times for each method to achieve specific target accuracies are provided in Table \ref{tab:comparison}. Notably, ComDML consistently demonstrates significant reductions in training time compared to the baselines, while preserving model accuracy across all scenarios. For example, ComDML achieves a remarkable 70\% reduction in training time compared to FedAvg and a substantial 71\% reduction compared to BrainTorrent on the I.I.D. CIFAR-10 dataset.
Unlike FedAvg, which can be hampered by communication delays with a central server, ComDML eliminates this bottleneck by enabling direct peer-to-peer communication. This not only accelerates model updates but also strengthens robustness against server failures and network disruptions, ultimately enhancing the overall learning process.

\subsubsection{Performance of ComDML with different numbers of agents}

To evaluate the scalability of ComDML, we assessed its performance across varying numbers of agents. Table \ref{tab:client_num} shows the training time of ComDML in comparison with baselines on the I.I.D. CIFAR-10 dataset, using different numbers of agents to achieve a target accuracy of 80\% for both ResNet-56 and ResNet-110 models. We employed a 20\% sampling rate for agent participation in each training round. As observed from Table \ref{tab:client_num}, increasing the number of agents does not negatively impact ComDML's performance, underscoring its robust scalability.

\begin{table}[t]
    \caption{Performance evaluation of training time (in seconds) with varying numbers of agents on I.I.D. CIFAR-10 dataset and comparison with other baselines for ResNet-56 and ResNet-110 models to reach a target accuracy of 80\%.}
    \label{tab:client_num}
    \centering
    \setlength{\tabcolsep}{2pt}
    \begin{tabular}{@{}c @{} c | c  c  c  c  c}
        \toprule
        \multirow{2}{*}{Model} & \multirow{2}{*}{Agents} & \multicolumn{5}{c}{Training Method} \\
        & & ComDML & Gossip L. & BrainTorrent & AllReduce & FedAvg\\
        \midrule
        \multirow{3}{*}{ResNet-56} & 20 & \textbf{7618} & 12637 & 14822 & 15660 & 14409\\
         & 50 & \textbf{9539} & 17716 & 20337 & 21339 & 19681\\
         & 100 & \textbf{10461} & 19465 & 22825 & 23652 & 22577 \\
         \midrule
        \multirow{3}{*}{ResNet-110} & 20 & \textbf{11799} & 18834  & 20234 & 19559 & 19322\\
         & 50 & \textbf{15014} & 25574 & 27753 & 28117 & 27191  \\
         & 100 & \textbf{15843} & 28825 & 31526 & 30085 & 29494\\
        \bottomrule
    \end{tabular}
\end{table}

\subsubsection{Integration of privacy protection methods}

ComDML smoothly integrates privacy-preserving methods with minimal overhead, effectively maintaining model accuracy. 
Our experiments on the CIFAR-10 dataset, employing ResNet-56 with 100 agents, demonstrate its ability to integrate privacy-preserving techniques without compromising either accuracy or training time. 
Remarkably, we obtained model accuracies of 81.7\% with distance correlation ($\alpha=0.5$) \cite{vepakomma2020nopeek}, 83.2\% with patch shuffling \cite{yao2022privacy}, and 77.6\% with differential privacy (using Laplace noise, $\epsilon=0.5$, $\delta=10^{-5}$) \cite{abadi2016deep} over experiments with 100 rounds (Other configuration details mirrored those in the referenced papers). These results compellingly showcase ComDML's flexibility in balancing privacy and performance without sacrificing efficiency.

\subsubsection{ComDML's performance with different network topologies}

ComDML's decentralized design, adaptable to diverse network structures from random to ring topologies \cite{beltran2023decentralized}, even copes with limited connections. It dynamically adjusts communication strategies for efficiency, outperforming centralized methods like FedAvg that are vulnerable to server bottlenecks.
By meticulously balancing communication overhead against computation, ComDML achieves faster overall training times compared to other baselines. It even adapts to extreme scenarios with poor links, allowing independent training if needed. This flexibility renders ComDML efficient across a wider range of network conditions than its counterparts.

\begin{figure}[t]
  \centering
   
   \includegraphics[width=1.00\linewidth]{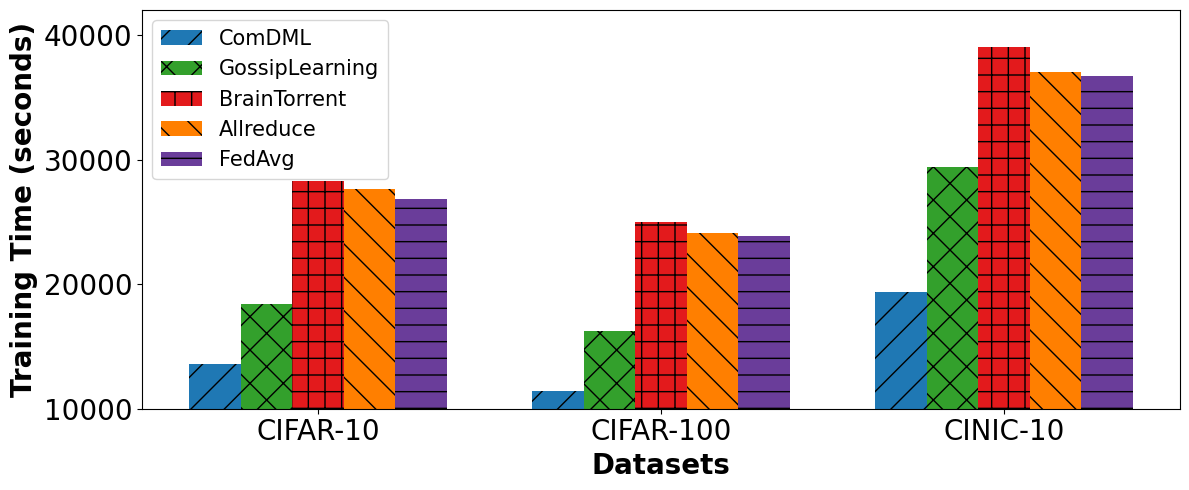}

   \caption{Comparison of total training time (in seconds) against baseline models under a 20\% link connectivity in random topology.}  
   \label{fig:com_link}
\end{figure}

Fig. \ref{fig:com_link} highlights ComDML's training efficiency compared to baselines under simulated limited communication links, where agents are randomly connected through only 20\% of the links present in a full graph. This setting mirrors those used in the 50-agent experiments across diverse I.I.D. datasets. ComDML's decentralized design maintains its efficiency even in scenarios with network disruptions or bandwidth limitations, ensuring persistent learning progress.

\section{Conclusion}
\label{sec:conclusion}

In this paper, we developed ComDML, an effective solution to address the challenges of collaboratively training large models in DML systems with diverse computational resources, communication bandwidth, and dataset sizes. This serverless framework enables slower agents to offload workloads to faster ones. 
By employing local-loss-based split learning, ComDML balances workloads, facilitating parallel updates and conquering resource constraints and straggler issues. 
We provided theoretical guarantees for the convergence of ComDML. Through extensive experiments on different datasets with heterogeneous agents, ComDML demonstrates remarkable reductions in training time without compromising model accuracy. It seamlessly integrates privacy-preserving techniques without sacrificing speed, ensuring both efficiency and confidentiality. Importantly, ComDML achieves these results without relying on a central server, thus offering a more scalable and efficient alternative to state-of-the-art DML methods.

\section*{Acknowledgment}

This work is supported in part by the National Science Foundation under Grants OIA-2148788, CAREER-2305491, CNS-2203239, CNS-2203412, and CCSS-2203238.

\section*{Appendix}
\label{sec:appendix}

\subsection{Proof of Theorem \ref{thm:slow_side}}

We show the convergence for both slow and fast agent sides, encompassing convex and non-convex functions. Slow agent-sides converge independently, while fast agent-sides rely on slow agent convergence for their own convergence, with rates presented.
Due to space, we show only the convergence for convex functions. The convergence for non-convex functions can be easily derived using the same techniques presented in the following.

First, we introduce a lemma that will be useful for proving the convergence of the fast agent-side later on.

\begin{lm}\label{lm:conv_C}   
Fubini's theorem, in conjunction with Assumption \ref{as:conv_pre_layer}, enables us to make the following observation:

\end{lm}

\begin{equation*}
{
\begin{aligned}
    \sum_{r} c^{a_s^m,{r}}&=\sum_{r} \int\left|d^{a_s^m,{r}}(\boldsymbol{z})-d^{a_s^m,{\star}}(\boldsymbol{z})\right|\\
    &=\int \sum_{r}\left|d^{a_s^m,{r}}(\boldsymbol{z})-d^{a_s^m,{\star}}(\boldsymbol{z})\right| d \boldsymbol{z}<\infty.
\end{aligned}
} 
\end{equation*}

Since $\sum_{r}\left|d^{a_s^m,{r}}(\boldsymbol{z})-d^{a_s^m,{\star}}(\boldsymbol{z})\right|$ is convergent, we have  $\left|d^{a_s^m,{r}}(\boldsymbol{z})-d^{a_s^m,{\star}}(\boldsymbol{z})\right| \rightarrow 0$. 

\subsubsection{Slow agent-side convergence} 

Assume that the slow agent-side functions satisfy Assumptions \ref{as:l_smooth}, \ref{as:mu_convex}, \ref{as:bound_variance}, and \ref{as:gradient_diss}. Considering the model update at round $r$, we have:

\begin{equation*}
{
\begin{aligned}
&\Delta \boldsymbol{w}^{{a_s^m}}=-\frac{\eta}{A^m} \sum_{i \in \mathcal{A}^m} g_k^{{a_s^m}}\left(\boldsymbol{w}_k^{{a_s^m}}\right)  \\
\Rightarrow& \mathbb{E}[\Delta \boldsymbol{w}^{{a_s^m}}]=-\frac{\eta}{K} \sum_{i} \mathbb{E}\left[\nabla f_k^{{a_s^m}}\left(\boldsymbol{w}_k^{{a_s^m}}\right)\right] .
\end{aligned}
}
\end{equation*}

We use the notation $\mathbb{E}$ to denote the expectation over all the randomness generated in the prior round. Building on the above, we separate the mean and variance by applying Lemma 4 of \cite{karimireddy2020scaffold}.

\begin{equation}\label{eq:proof_after_sep_var_mean}
\begin{aligned}
&\mathbb{E}\left\|\boldsymbol{w}^{a_s^m}+\Delta \boldsymbol{w}^{a_s^m}-\boldsymbol{w}^{a_s^m\star}\right\|^2 \leq \\
&\left\|\boldsymbol{w}^{a_s^m}-\boldsymbol{w}^{a_s^m\star}\right\|^2 \\ & \underbrace{-\frac{2 \eta}{K} \sum_{i}\left\langle\nabla f_k^{a_s^m}\left(\boldsymbol{w}_k^{a_s^m}\right), \boldsymbol{w}^{a_s^m}-\boldsymbol{w}^{a_s^m\star}\right\rangle}_{\mathcal{T}_1} \\
& +\underbrace{\eta^2 \mathbb{E}\left\|\frac{1}{A^{m,r}} \sum_{i \in \mathcal{A}^{m,r}} \nabla f_k^{a_s^m}\left(\boldsymbol{w}_k^{a_s^m}\right)\right\|^2}_{\mathcal{T}_2}+\frac{\eta^2 \sigma^2}{A^{m,r}}.
\end{aligned}
\end{equation}

By applying Lemma 5 of \cite{karimireddy2020scaffold} to $\mathcal{T}_1$, it is observed that:
 
\[
\begin{aligned}
& \mathcal{T}_1  \leq \\ & -2 {\eta}\left(f^{a_s^m}(\boldsymbol{w}^{a_s^m})-f^{a_s^m}\left(\boldsymbol{w}^{a_s^m\star}\right)+\frac{\mu}{4}\left\|\boldsymbol{w}^{a_s^m}-\boldsymbol{w}^{a_s^m\star}\right\|^2\right).
\end{aligned}
\]

The bound on $\mathcal{T}_2$ is obtained by combining the relaxed triangle inequality, Lemma 3 of \cite{karimireddy2020scaffold}, and Assumption \ref{as:gradient_diss} as follows:

\[
\begin{aligned}
\mathcal{T}_2 \leq
&8 \eta^2 L\left(B^2+1\right)\left(f^{a_s^m}(\boldsymbol{w}^{a_s^m})-f^{a_s^m}\left(\boldsymbol{w}^{a_s^m{\star}}\right)\right)\\
&+\left(1-\frac{A^{m,r}}{K}\right) \frac{4 {\eta}^2}{A^{m,r}} G_2^2 .
\end{aligned}
\]

After plugging back the bounds $\mathcal{T}_1$ and $\mathcal{T}_2$ into (\ref{eq:proof_after_sep_var_mean}), we rearrange and move the expression $(f^{a_s^m}(\boldsymbol{w}^{a_s^m}) - f^{a_s^m}(\boldsymbol{w}^{a_s^m{\star}}))$ and then divide throughout by $\eta$, while assuming $4 L \eta (B^2 + 1) \leq 1$.

\begin{equation*}\label{eq:proof_last_step}
\begin{aligned}
&f^{a_s^m}(\boldsymbol{w}^{a_s^m,{r}})-f^{a_s^m}(\boldsymbol{w}^{a_s^m,{\star}}) \leq \\ & \frac{1}{\eta} 
  \left(1 - \frac{\eta \mu}{4}\right)\left\|\boldsymbol{w}^{a_s^m,{r}}-  \boldsymbol{w}^{a_s^m,{\star}}\right\|^2 \\
& - \frac{1}{\eta} \left\|\boldsymbol{w}^{a_s^m,{r+1}}-\boldsymbol{w}^{a_s^m,{\star}}\right\|^2 \\ &
+ \eta \left( \frac{\sigma^2}{A^{m,r}}+\frac{4G_2^2}{A^{m,r}} \left(1-\frac{A^{m,r}}{K}\right) \right). \\
\end{aligned}
\end{equation*}

Applying the linear convergence rate lemma (Lemma 1 of \cite{karimireddy2020scaffold}), we obtain the desired convergence rate for the convex case.

\[
{
\begin{aligned}
&\mathbb{E}\left[f^{a_s^m}\left({\overline{\boldsymbol{w}^{a_s^m}}^R}\right)\right]-f^{a_s^m}\left(\boldsymbol{w}^{a_s^m{\star}}\right) \\
&= \mathcal{O}\left(\frac{\eta H_1^2 }{\mu{R A^{m,r}}}+\mu D^2\exp\left(-\frac{\mu}{L(1+B^2)}{R}\right)\right),
\end{aligned}}
\]
where $H_1^2:=\sigma^2+\left(1-\frac{A^m}{K}\right) G_2^2$, $D:=\left\|\boldsymbol{w}^{{a_s^m}^0}-\boldsymbol{w}^{{a_s^m}^{\star}}\right\|$,  and $A^{m} = \min_r \{ A^{m,r} > 0\}$. 

The proof for the non-convex case follows a similar approach, with the key difference that Assumption~\ref{as:mu_convex} is disregarded. It relies on the sub-linear convergence rate lemma (Lemma 2 of \cite{karimireddy2020scaffold}). This leads to:

\[
{ 
\begin{aligned}
&\mathbb{E}\left[\left\|\nabla f^{a_s^m}\left(\overline{\boldsymbol{w}^{a_s^m}}^R\right)\right\|^2\right] = \mathcal{O}\left(\frac{L H_1 \sqrt{F^{a_s^m}}}{\sqrt{R {A^{m,r}}}}+\frac{B^2 L F^{a_s^m}}{R}\right),
\end{aligned}}
\]
where $F^{a_s^m}:=f^{a_s^m}\left(\boldsymbol{w}^{a_s^m0}\right)-f^{a_s^m{\star}}$. 

\subsubsection{Fast agent-side convergence} Now we provide convergence analysis of fast agent-side models. Assume that the fast agent-side functions satisfy Assumptions \ref{as:l_smooth}, \ref{as:mu_convex}, \ref{as:bound_variance}, and \ref{as:gradient_diss}. The model's update adheres to the following equation:

\begin{equation*}
{
\begin{aligned}
&\Delta \boldsymbol{w}^{{a_f^m}}=-\frac{\eta}{A^{m,r}} \sum_{i \in \mathcal{A}^{m,r}} g_k^{{a_f^m}}\left(\boldsymbol{w}_k^{{a_f^m}}\right)  \\
\Rightarrow& \mathbb{E}[\Delta \boldsymbol{w}^{{a_f^m}}]=-\frac{\eta}{K} \sum_{i} \mathbb{E}\left[\nabla f_k^{{a_f^m}}\left(\boldsymbol{w}_k^{{a_f^m}}\right)\right] .
\end{aligned}
}
\end{equation*}

By quadratic upper bound of Assumption \ref{as:l_smooth} (L-smoothness), we have:

\[
{
\begin{aligned}
  &f^{{a_f^m}}( 
  \boldsymbol{w}^{a_f^m,{r+1}})\leq  f^{{a_f^m}}(\boldsymbol{w}^{a_f^m,{r}}) +\frac{L}{2}\| \boldsymbol{w}^{a_f^m,{r+1}}-\boldsymbol{w}^{a_f^m,{r}}\|^2 \\
  &+   \nabla f^{{a_f^m}}(\boldsymbol{w}^{a_f^m,{r}})^T(\boldsymbol{w}^{a_f^m,{r+1}} - \boldsymbol{w}^{a_f^m,{r}}) 
  .
\end{aligned}
}
\]

We incorporate the weight update formula into the above inequality, followed by taking expectation across all randomness. This process yields the following:

\[
{
\begin{aligned}
    &\mathbb{E} \left[f^{{a_f^m}}(\boldsymbol{w}^{a_f^m,{r+1}})\right] \leq 
    \mathbb{E} \left[f^{{a_f^m}}(\boldsymbol{w}^{a_f^m{r}})\right] \\ 
    &{ \underbrace{-\eta\mathbb{E} \left[\nabla f^{{a_f^m}}(\boldsymbol{w}^{a_f^m,{r}})^T\left(\frac{1}{{A}^{m,r}} \sum_{ \boldsymbol{A}^{m,r}} \nabla f^{{a_f^m}}_k\left(\boldsymbol{u}^{a_s^m,{r}} ; \boldsymbol{w}^{a_f^m,{r}}\right)\right)\right]}_{\mathcal{T}_3}} \\ 
    &+\frac{L}{2}\underbrace{ \eta^2 \mathbb{E}  {\left[\left\|\frac{1}{{A}^{m,r}} \sum_{\boldsymbol{A}^{m,r}} \nabla f^{{a_f^m}}_k\left(\boldsymbol{u}^{a_s^m,{r}} ; \boldsymbol{w}^{a_f^m,{r}}\right)\right\|^2\right]}}_{\mathcal{T}_4}.
\end{aligned}
}
\]

We now prove the boundedness of $\mathcal{T}_3$ and $\mathcal{T}_4$. 
By applying Cauchy-Schwartz and Jensen's inequality, and leveraging Assumption \ref{as:mu_convex} along with Lemma \ref{lm:conv_C}, we obtain the following bound for $\mathcal{T}_3$:

\begin{equation*}
{
\begin{aligned} \label{eq:15}
\mathcal{T}_3 
& \leq \eta \sqrt{2 G_1^2 (G_2^2+2LB^2( f^{a_f^m}(\boldsymbol{w}^{a_f^{m,r}}) -f^{a_f^m{\star}}))c^{a_s^m,{r}}} \\
&  -\eta \mathbb{E}\left[\left\|\nabla f^{{a_f^m}}(\boldsymbol{w}^{a_f^{m,r}})\right\|^2\right]. 
\end{aligned}
}
\end{equation*}

Leveraging Assumption \ref{as:mu_convex} and mirroring the approach for the slow agent-side function, we establish the bound for $\mathcal{T}_4$:

\[
{
\begin{aligned}
     \mathcal{T}_4 \leq &
     8 \eta^2 L\left(B^2+1\right)\left(f_k^{a_f^m}( \boldsymbol{w}^{a_f^m})-f_k^{a_f^m}\left( \boldsymbol{w}^{a_f^m{\star}}\right)\right) \\
     & +\left(1-\frac{A^{m,r}}{K}\right) \frac{4 {\eta}^2}{A^{m,r}} G_2^2.
\end{aligned}
}
\]

By using the bounds of $\mathcal{T}_3$ and $\mathcal{T}_4$ and assuming $4 L \eta (B^2 + 1) \leq 1$, we have:

{
\begin{equation*}
\begin{aligned}
\label{eq:inequality}
&\mathbb{E}\left[\left\|\nabla f^{{a_f^m}}(\boldsymbol{w}^{a_f^{m,r}})\right\|^2\right]  \leq  \frac{\mathbb{E}\left[f^{{a_f^m}}(\boldsymbol{w}^{a_f^{m,r}})\right]}{\eta} \\ & - \frac{\mathbb{E}\left[f^{{a_f^m}}(\boldsymbol{w}^{{s_f^m,r+1}})\right]}{\eta}\\
&+2 \eta L^3\left(B^2+1\right)\left(f^{{a_f^m}}(\boldsymbol{w}^{{a_f^m0}})-f^{{a_f^m{\star}}}\right)  \\ & + \left(1-\frac{A^m}{K}\right) \frac{{\eta}L^2}{A^m} G_2^2 \\
&+\sqrt{2 G_1^2 (G_2^2+2LB^2( f^{a_f^m}(\boldsymbol{w}^{a_f^{m,r}}) -f^{s_f^m{\star}})) c^{a_s^m,{r}}}.\\
\end{aligned}
\end{equation*} }

By applying Lemma \ref{lm:conv_C}, Lemma 2 from \cite{karimireddy2020scaffold}, defining $F^{{a_f^m}}:= \max_r \{f^{{a_f^m}}\left(\boldsymbol{w}^{a_f^{m,r}}\right)-f^{a_f^m{\star}} \}$, and averaging the summation over the third term, we obtain:

\[
{ 
\begin{aligned}
\mathbb{E}\left[\left\|\nabla f^{{a_f^m}}(\boldsymbol{w}^{a_f^m,{r}})\right\|^2\right]  = 
\mathcal{O}\left(\frac{C_1}{R} + \frac{H_2 \sqrt{F^{a_f^m{0}}} }{\sqrt{R A^{m,r}}} + \frac{ F^{a_f^m{0}}}{\eta R}  \right),
\end{aligned}}
\]
where $H_2^2:= L^3\left(B^2+1\right)F^{a_f^m{0}}+\left(1-\frac{A^m}{K}\right) {L^2G_2^2}$, $F^{a_f^m}:=f^{a_f^m}\left(\boldsymbol{w}^{a_f^m0}\right)-f^{a_f^m{\star}}$, and $C_1 = G_1\sqrt{G_2^2+2LB^2F^{a_f^m{0}}\sum_rc^{a_s^m,{r}}}$.

The proof for the non-convex case follows a similar ap-
approach by disregarding Assumption \ref{as:mu_convex}. Using telescoping sum, the proof for non-convex functions yields the following rate.
\[
{ 
\begin{aligned}
\mathbb{E}\left[\left\|\nabla f^{{a_f^m}}(\boldsymbol{w}^{a_f^m,{r}})\right\|^2\right]  = 
\mathcal{O} \left( \frac{C_2}{R}  + \frac{H_2\sqrt{{F^{a_f^m{0}}} }}{\sqrt{R{A^{m,r}}}}  + \frac{{F^{a_f^m{0}}}}{\eta R}\right),
\end{aligned}}
\]
where $C_2 = G_1\sqrt{G_2^2+B^2G_1^2\sum_rc^{a_s^m,{r}}}$. The fast agent-side bound has an extra term due to its dependence on the slow agent-side model convergence, leading to a looser bound.

\bibliographystyle{IEEEtran}
\bibliography{egbib.bib}

\end{document}